\documentclass[letterpaper, 10 pt, conference]{ieeeconf}  

\IEEEoverridecommandlockouts                              

\usepackage{amsmath,amsfonts}
\usepackage{algorithm}
\usepackage{algpseudocode}
\usepackage{array}
\usepackage[caption=false,font=normalsize,labelfont=sf,textfont=sf]{subfig}
\usepackage{makecell}
\usepackage{hyperref}
\usepackage{cleveref}  
\usepackage{amssymb}
\usepackage{svg}
\usepackage{gensymb}
\usepackage{textcomp}
\usepackage{booktabs}
\usepackage{stfloats}
\usepackage{url}
\usepackage{verbatim}
\usepackage{graphicx}
\usepackage{multirow}
\usepackage{soul}
\usepackage{newtxtext, newtxmath} 
\usepackage{xspace}
\usepackage{fontawesome5}

\usepackage[dvipsnames]{xcolor}
\usepackage[table]{xcolor}

\usepackage{dsfont}              

\usepackage{color}
\usepackage{capt-of}
\definecolor{highlight}{RGB}{205, 232, 248}
\definecolor{lightpurple}{RGB}{247, 228, 252}

\newcommand{\eg}{\textit{e.g.}\xspace}

\title{\textbf{Continuous Actions from Discrete Minds: Latent-Aligned Planning for End-to-End Autonomous Driving}}

\author{Ruoyu Yao$^{1,*}$, Yusen Xie$^{1,*}$, Qingzhao Liu$^{1}$, Pei Liu$^{1}$, Zewei Yang$^{1}$, Yipeng Zhu$^{2}$, \\ Xiaolong Wang$^{2,\dagger}$, Jun Ma$^{1,\text{\faEnvelope}}$\, \textit{Senior Member, IEEE}
\thanks{1. The Hong Kong University of Science and Technology (Guangzhou), Guangzhou 511453, China (e-mail: jun.ma@ust.hk); 
        2. Central Media Technology Institute, Huawei (e-mail: wangxiaolong26@huawei.com); $*$: Equal contribution; $\dagger$: Project lead; \faEnvelope: Corresponding author.} 
}

\begin{document}
\maketitle

\begin{abstract}
Bridging the gap between the discrete reasoning of Vision-Language Models and the continuous, physics-constrained nature of autonomous driving remains a significant challenge. 
In this work, we introduce LaPla, a unified Vision-Language-Action (VLA) framework featuring latent-aligned planning to seamlessly ground semantic understanding in precise motion execution. 
We first design an action tokenizer based on a residual vector-quantized variational autoencoder (VQ-VAE), capturing vehicle kinematics and encoding trajectory features into a structured latent space.
Rather than discrete codebook lookups that inevitably introduce quantization errors, LaPla repurposes this representation as a physical prior to bridge the modality gap between high-dimensional semantics and the raw action space.
Specifically, given multimodal inputs integrating multi-view images, historical actions, and textual instructions, LaPla incorporates concurrent action queries to causally attend to the multimodal context in a single forward pass, projecting hidden states directly into the pretrained VQ-VAE latent space. 
The frozen decoder then translates these continuous latents into actions, effectively eliminating quantization errors and ensuring physically plausible trajectories while bypassing time-consuming autoregressive generation.
Extensive experiments on the nuScenes benchmark demonstrate that LaPla achieves competitive open-loop performance, reducing long-horizon L2 error by 15.52\,\% compared to state-of-the-art VLA methods. 
Closed-loop evaluations on the NVIDIA AlpaSim simulator further confirm its superior capability in ensuring smooth driving progress, improving the success rate by 33.34 percentage points with significantly reduced inference latency.
\end{abstract}

\section{Introduction}
The pursuit of fully autonomous driving (AD) has witnessed a paradigm shift from modular pipelines to end-to-end learning architectures. Recently, the advent of Large Language Models (LLMs) and Vision-Language Models (VLMs) has revolutionized this landscape, catalyzing the evolution of Vision-Language-Action (VLA) models~\cite{jiang2025survey}. Benefiting from the synergy between pretrained world knowledge and domain-specific driving policies, these VLA models exhibit unprecedented capabilities in complex scene understanding~\cite{zhou2026opendrivevla}, interpretable decision-making~\cite{xu2024drivegpt4}, and predictive reasoning~\cite{wang2025omnidrive}. Consequently, equipping AD with unified VLA frameworks has become a prevailing trend, promising to tackle long-tail open-world driving scenarios that have traditionally bottlenecked conventional systems.

\begin{figure}[t] 
\centering 
\includegraphics[width=\linewidth]{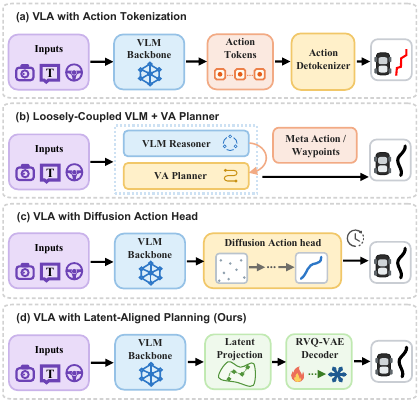} 
\caption{
\textbf{A comparison between existing VLA architectures and our approach.} Action tokenization approaches suffer from quantization errors in trajectory generation. 
Dual-system designs incur a disconnect between high-level semantics and low-level planning, while diffusion-based action heads introduce computational latency from iterative denoising. 
In contrast, our architecture plans directly in a kinematically sound latent action space, enabling both efficient and smooth trajectory generation.
}
\label{fig:architec compare}
\vspace{-1.5em}
\end{figure}

Despite these advanced cognitive capabilities, grounding the high-dimensional semantic representations of such large foundation models into the continuous, physics-constrained control required for AD remains a fundamental bottleneck.
Current literature explores this alignment through three primary avenues. 
The first explicitly forces continuous trajectories into the language modeling paradigm via action tokenization, enabling unified autoregressive architectures~\cite{wu2024smart, zhou2026autovla}. Although conceptually straightforward, the discretization inevitably introduces quantization errors. Consequently, these frameworks may attain favorable performance in open-loop evaluations but suffer from error amplification in closed-loop scenarios, resulting in jagged and physically unsafe maneuvers. 
The second approach bypasses action tokenization by employing loosely coupled dual-system architectures, cascading an upper-level VLM into a downstream planner~\cite{jiang2024senna, tian2024drivevlm}. However, this fragmented design inherently downgrades rich multimodal reasoning into coarse decisions and necessitates asynchronous execution, introducing a fundamental spatio-temporal disconnect between high-level intents and reactive planning.
The third avenue bridges VLM backbones and generative action heads represented by diffusion models~\cite{liu2026drivepi}, ensuring smooth trajectories via the probabilistic noise-denoising process in continuous spaces. However, learning kinematic features from scratch with diffusion action heads is nontrivial, which leads to training instability and knowledge erosion when jointly optimized with the VLM backbone~\cite{driess2026knowledge}. Moreover, the iterative denoising process incurs prohibitive computational latency, rendering it sub-optimal for strict real-time deployment on resource-constrained vehicles.

To overcome the limitations of planning directly in the raw action space, a promising strategy shifts the action representation to planning within learned latent spaces~\cite{zheng2024genad, zhang2025lap}. 
By learning a structured, non-linear manifold of plausible driving behaviors, this approach abstracts complex motion correlations into a unified embedding space, allowing VLMs to align their high-dimensional semantic reasoning with a safe, continuous physical region~\cite{fu2025orion}.
Recent approaches mainly pursue this direction by integrating physics priors into latent reasoning~\cite{luo2026last, lu2026onevl} or unifying world modeling with planning in a shared latent space~\cite{liu2026driveworld, lu2026dawn}.
A common thread across these methods is that the latent space is co-learned with the entire VLA architecture, such that the action representation and the decoder are jointly optimized from scratch.
We argue that this co-learning formulation, despite its flexibility, imposes an excessive optimization burden on the VLA by compounding representation discovery with policy learning. Bereft of a stable prior, the VLA is forced to concurrently establish the topological coordinate system and solve the driving policy within it.
A more tractable path requires anchoring the VLA with a frozen, kinematically sound latent structure, transitioning the model's objective from discovering a shifting space to solely executing policies within an invariant one.

To fill this critical void, we propose \textbf{LaPla}, a unified VLA framework featuring \textbf{La}tent-aligned \textbf{Pla}nning to elegantly bridge the discrete semantic reasoning of VLMs with the continuous, kinematic-constrained actions of AD. 
Diverging from both holistic action tokenization and intricate curriculum learning for semantic-to-action alignment, LaPla performs end-to-end planning directly within a well-structured latent manifold. 
Specifically, we pretrain a residual vector-quantized variational autoencoder (VQ-VAE)~\cite{zeghidour2021soundstream} on large-scale trajectory data to capture vehicle kinematics and motion patterns. 
During pretraining, the discrete bottleneck of the VQ-VAE is essential to cluster the infinite trajectory space into distinct, physically plausible primitives.
During VLA planning, rather than performing hard, discrete codebook lookups that introduce quantization errors, LaPla maps rich multimodal contextual inputs directly into the continuous embedding space spanned by these learned primitives. 
By feeding these continuous latents into the frozen decoder, we bypass quantization bottlenecks to ensure smooth motion execution.
Consequently, this mechanism yields robust driving stability over extended planning horizons in closed-loop environments.
A comparison between our approach and existing VLA architectures is shown in Fig.~\ref{fig:architec compare}. We summarize our contributions as follows:

\begin{itemize}
    \item We propose LaPla, a unified VLA framework that formulates autonomous driving as a latent-aligned sequence modeling task. It seamlessly integrates multimodal context into high-precision trajectory generation, fundamentally bridging the mismatch between discrete semantic reasoning and continuous planning tasks.

    \item We pioneer an implicit latent-aligned planning mechanism governed by a frozen residual VQ-VAE decoder. The frozen decoder serves as a kinematic gradient preconditioner that projects raw Euclidean errors into physically consistent gradient updates to actively steer the VLM’s optimization, thereby bypassing quantization bottlenecks while steering the output toward the kinematically plausible manifold.

    \item We conduct extensive experiments to validate our framework. LaPla achieves competitive open-loop performance on nuScenes~\cite{caesar2020nuscenes}, demonstrating physically coherent end-to-end planning that mitigates long-horizon drift observed in state-of-the-art methods. Closed-loop evaluations on the AlpaSim simulator~\cite{alpasim_2025} further demonstrate its significant advancements in both success rate and inference speed.
    
\end{itemize}

\begin{figure*}
    \centering
    \includegraphics[width=\linewidth]{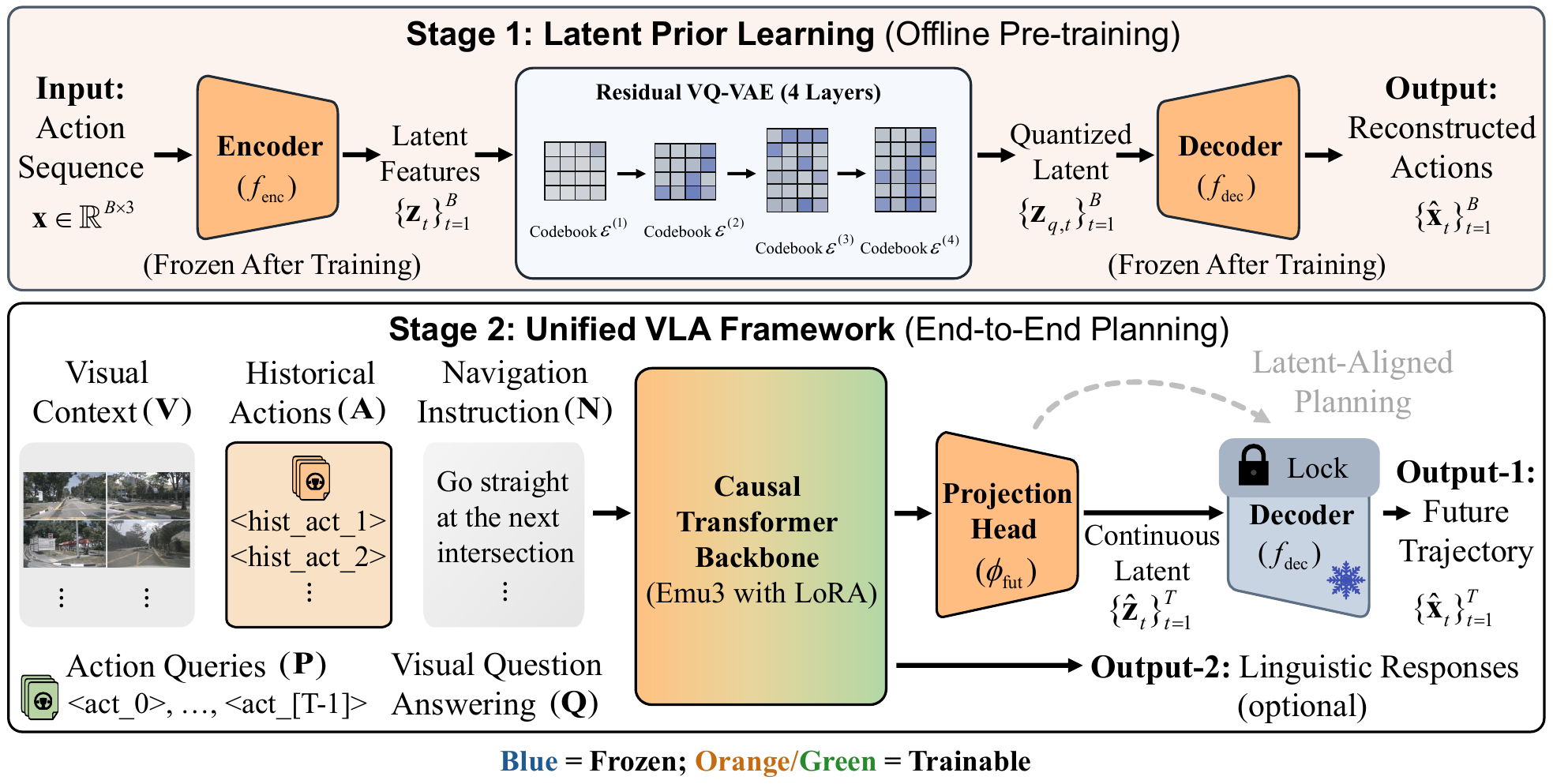}
    \caption{\textbf{Overall framework of LaPla.} \textbf{Top:} A pretrained residual VQ-VAE yields a frozen decoder $f_{\text{dec}}$ serving as a trajectory prior. \textbf{Bottom:} The VLA model ingests multimodal contexts and predicts continuous future latents $\{\hat{\mathbf{z}}_t\}^T_{t=1}$. Guided by a Smooth L1 loss, $f_{\text{dec}}$ decodes $\hat{\mathbf{z}}_t$ into future action trajectory $\{\hat{\mathbf{x}}_t\}^T_{t=1}$, achieving end-to-end latent-aligned planning.}
    \vspace{-1.5em}
 
    \label{fig:pipeline}
\end{figure*}

\section{Related Work}

\subsection{End-to-End Autonomous Driving}
Traditional AD systems adopt a modular pipeline decomposing the driving task into perception, prediction, planning, and control components. While the modularity enables decoupled development of individual components, it suffers from error accumulation across modules, inconsistent optimization objectives, and limited scalability to long-tail scenarios~\cite{chen2024end}.
End-to-end AD offers an alternative employing differentiable frameworks that map raw sensor inputs directly to continuous vehicle trajectories. Classic end-to-end approaches rely primarily on Vision-to-Action (VA) architectures, leveraging imitation learning or reinforcement learning to achieve holistic planning-oriented optimization~\cite{hu2022st, hu2023planning, liu2026reinforced}. Nevertheless, without a linguistic modality, VA models offer no principled mechanism for inspecting their internal reasoning, steering their behavior with explicit commands, or grounding their decisions in rich, open-world knowledge~\cite{dong2025end}.
More recently, foundation models have further catalyzed end-to-end systems with stronger generalization and enhanced reasoning capabilities~\cite{wang2025omnidrive}. This evolution naturally leads to the emergence of VLA models that reconcile perception, semantic understanding, and action generation within a comprehensive framework.

\subsection{Vision-Language-Action Models for Autonomous Driving}
Existing VLA models can be categorized into unified and dual-system paradigms based on how they organize reasoning and execution~\cite{hu2025vision}.
Unified methods integrate semantic reasoning and action generation within a single model. Early studies build comprehensive frameworks to jointly address perception, reasoning, and planning, outputting trajectories directly as language tokens~\cite{hwang2024emma, wang2025omnidrive, zhou2026opendrivevla}. To enhance generation capability, AutoVLA employs action tokenization for dedicated representation learning~\cite{zhou2026autovla}, while other works typically adopt generative action heads for smooth continuous planning~\cite{liu2026drivepi, huang2026mindvla}.
In contrast, dual-system methods decouple high-level reasoning from low-level execution in loosely coupled frameworks. DriveVLM introduces a slow-fast dual system where a VLM handles Chain-of-Thought (CoT) reasoning and a VA refines high-frequency control~\cite{tian2024drivevlm}. Senna cascades a VLM-based discrete decision-maker with a VA planner to ground commonsense reasoning~\cite{jiang2024senna}, and Senna-2 incorporates a decision adapter to mitigate misalignment~\cite{song2026senna}.
Despite their progress, both architectures face a persistent challenge: action generation lacks structural physical grounding. This motivates a shift from directly regressing raw waypoints to planning within learned, kinematically sound latent action spaces.

\section{Methodology}
Our methodology is structured in two stages, as shown in Fig.~\ref{fig:pipeline}. We first pretrain a trajectory VQ-VAE and freeze its decoder to serve as a \textbf{frozen latent prior} that encapsulates physically plausible motion patterns. Building upon this prior work, we then introduce \textbf{LaPla}, a unified VLA framework that predicts continuous latent vectors to directly drive the frozen decoder for trajectory generation, while jointly performing visual question answering via shared multimodal context. This decomposition cleanly separates the acquisition of the generic trajectory prior from the policy learning, enabling fully end-to-end differentiable planning without resorting to discrete token generation.

\subsection{Latent Prior Learning}\label{sec: frozen prior}
Directly regressing waypoints from scratch forces the model to re-discover basic vehicle kinematics and multi-modal maneuver structures from limited driving data. To relieve this burden, we first construct a generative module that projects the space of plausible trajectories into a structured latent manifold.  Once frozen, its decoder provides a strong prior that biases the output toward the region of smooth driving patterns observed in pretraining, thereby relieving the downstream policy from learning kinematic regularities from scratch.

Specifically, we pretrain a residual VQ-VAE to serve as a frozen latent prior for the VLA framework.
A trajectory is represented as a sequence of local displacement and heading features $\mathbf{x} \in \mathbb{R}^{B \times 3}$, obtained by differencing raw waypoints and computing frame-wise orientation angles.
We build an encoder network \(f_{\text{enc}}\) to map the action at each timestep to its latent features, incorporating historical motion context. A shared MLP \(g^{(1)}_{\text{act}}\) transforms each action into a temporal motion embedding, followed by pooling over the local time window and concatenation of the pooled representation with the current-step embedding. Another shared MLP \(g^{(2)}_{\text{act}}\) then encodes the obtained vectors and generates the latent action features:
\begin{equation}
\begin{aligned}
    \mathbf{m}_t &= g^{(1)}_{\text{act}}(\mathbf{x}_t), \\
    \mathbf{c}_t &= \frac{1}{\omega} \sum_{\tau = t-\omega+1}^{t} \mathbf{m}_\tau, \\
    \mathbf{z}_{t} &= g^{(2)}_{\text{act}}([\mathbf{c}_t; \mathbf{m}_t]),
\end{aligned}
\end{equation}
where \(\mathbf{m}_t\) denotes the temporal motion feature at each timestep. \(\omega\) is the window size, and \(\mathbf{c}_t\) denotes the mean-pooled feature over the past \(\omega\) steps. \([\cdot;\cdot]\) represents concatenation along the feature dimension. The resulting \(\mathbf{z}_t\) thus carries both the instantaneous motion at step \(t\) and the local trend of the recent trajectory, which is then quantized through $L = 4$ residual layers with codebooks $\mathcal{E}^{(l)} = \{\mathbf{e}_k^{(l)}\}_{k=1}^{K_l}$ ($K_l \in \{64, 64, 128, 128\}$):
\begin{equation}
\begin{aligned}
\mathbf{r}_t^{(0)} &= \mathbf{z}_t, \\
i_t^{(l)} &= \arg\min_k \big\|\mathbf{r}_t^{(l-1)} - \mathbf{e}_k^{(l)}\big\|_2, \\
\mathbf{r}_t^{(l)} &= \mathbf{r}_t^{(l-1)} - \mathbf{e}_{i_t^{(l)}}^{(l)},
\end{aligned}
\end{equation}
where $\mathbf{r}_t^{(l)}$ is the residual after $l$ layers, $i_t^{(l)}$ the selected code index, and $\mathbf{e}_{i_t^{(l)}}^{(l)}$ the corresponding codebook vector.
The quantized latent is the sum $\mathbf{z}_{q,t} = \sum_{l=1}^{L} \mathbf{e}_{i_t^{(l)}}^{(l)}$, and a decoder $f_{\text{dec}}$ reconstructs the motion features $\hat{\mathbf{x}}_t = f_{\text{dec}}(\mathbf{z}_{q,t})$.
Training minimizes the combined reconstruction and commitment loss:
\begin{equation}
\mathcal{L}_{\text{prior}} = \frac{1}{B}\sum^B_{t=1} \big\|\mathbf{x}_t - \hat{\mathbf{x}}_t\big\|_2^2 + \sum_{l=1}^{L} \big\|\mathbf{r}_t^{(l-1)} - \text{sg}[\mathbf{e}_{i_t^{(l)}}^{(l)}]\big\|_2^2,
\end{equation}
where $\mathrm{sg}[\cdot]$ denotes the stop-gradient operator.
The codebooks themselves are updated via an exponential moving average with dead-code revival following standard practice.

This coarse-to-fine residual design imposes a hierarchical decomposition of maneuvers: early layers capture macroscopic motion primitives (\eg, turning vs.\ straight), while deeper layers add fine-grained corrections.
As a result, the latent space becomes densely populated and combinatorially expressive (\(\prod K_l \approx 6.7\times10^7\) possible codes), while remaining semantically smooth since nearby latents correspond to similar trajectories, greatly facilitating the subsequent latent alignment learning in LaPla.

\subsection{Unified VLA Framework with Latent-Aligned Planning}\label{sec: vla framework}
\label{sec:vla_framework}

Given the frozen trajectory prior \(f_{\text{dec}}\), LaPla formulates end-to-end driving as a multimodal sequence modeling problem. The model jointly predicts a future trajectory via latent-aligned planning and answers driving-related questions through an auxiliary vision-question-answering (VQA) task, all within a single causal transformer backbone.

\subsubsection{Multimodal Sequence Formulation}
\label{sec:multimodal_sequence}
We employ Emu3-8B~\cite{wang2024emu3} as the VLM backbone to construct our framework, benefiting from its exceptional performance in multimodal understanding and generation tasks. 
LaPla constructs a single token sequence that integrates visual, action, and textual modalities. The full input sequence is
\begin{equation}
\mathbf{X} = \big[ \mathbf{V}_{-H}, \dots, \mathbf{V}_0, \; \mathbf{A}_{-H}, \dots, \mathbf{A}_{-1}, \; \mathbf{N}, \; \mathbf{P}, \; \mathbf{Q} \big],
\label{eq:input_sequence}
\end{equation}
where \(H\) is the number of historical steps and time indices \(t \le 0\) denote steps relative to the current planning moment. Each component is defined as follows.

\textbf{Visual Context.} The visual tokens \(\mathbf{V}_t\) at historical step \(t\) correspond to the multi-view camera images captured at that instant. Images from the front, front-left, front-right, and back cameras are partitioned into patches and encoded into visual token sequences by Emu3's visual tokenizer. The prefix \(\mathbf{V}_{-H}, \dots, \mathbf{V}_0\) thus provides a sequential observation containing \(H\) historical frames and the current frame.

\textbf{Historical Actions.} The ego-motion history is encoded in parallel with the visual stream. At each past step \(t\), the displacement and heading feature \(\mathbf{x}_t\) is passed through the frozen VQ-VAE encoder \(f_{\text{enc}}\) and the residual quantization layers to obtain the quantized latent vector \(\mathbf{z}_{q,t}\). An MLP then projects this latent into the shared embedding space, yielding the history token \(\mathbf{A}_t = \phi_{\text{hist}}(\mathbf{z}_{q,t})\), which abstracts the recent motion into a semantically aligned representation.

\textbf{Navigation Instruction.} The navigation instruction combines a high-level directional command (\eg, ``turn left,'' ``go straight'') with a structured planning task description that explicitly enforces collision avoidance, traffic rule compliance, and kinematic feasibility. This textual directive is tokenized by Emu3's text tokenizer into the sequence \(\mathbf{N} = (n_1, \dots, n_{L_N})\).

\textbf{Learnable Action Queries and VQA.} After the context, we insert a sequence of \(T\) special learnable queries \(\mathbf{P} = (\texttt{<act\_0>}, \dots, \texttt{<act\_[T-1]>})\) for the future trajectory. An optional VQA suffix \(\mathbf{Q}\) (only used during training) is appended beyond these placeholders and is described in \Cref{sec:vqa_auxiliary}.

This ordering allows the model to progressively build a grounded scene representation before being guided by the navigation intent. The entire sequence is processed by a causal transformer. With the standard causal mask, the attention operation at position \(i\) is
\begin{equation}
\mathbf{o}_i = \sum_{j=1}^{i} \alpha_{ij} \mathbf{v}_j, \quad
\alpha_{ij} = \frac{\exp(\mathbf{q}_i^\top \mathbf{k}_j / \sqrt{d_k})}{\sum_{u=1}^{i} \exp(\mathbf{q}_i^\top \mathbf{k}_u / \sqrt{d_k})},
\label{eq:causal_attention}
\end{equation}
where \(\mathbf{q}_i\) is the query vector at position \(i\), \(\mathbf{k}_j\) and \(\mathbf{v}_j\) are the key and value vectors at position \(j\), and \(d_k\) is the dimensionality of the key vectors.

\begin{figure}[t] 
\centering 
\includegraphics[width=\linewidth]{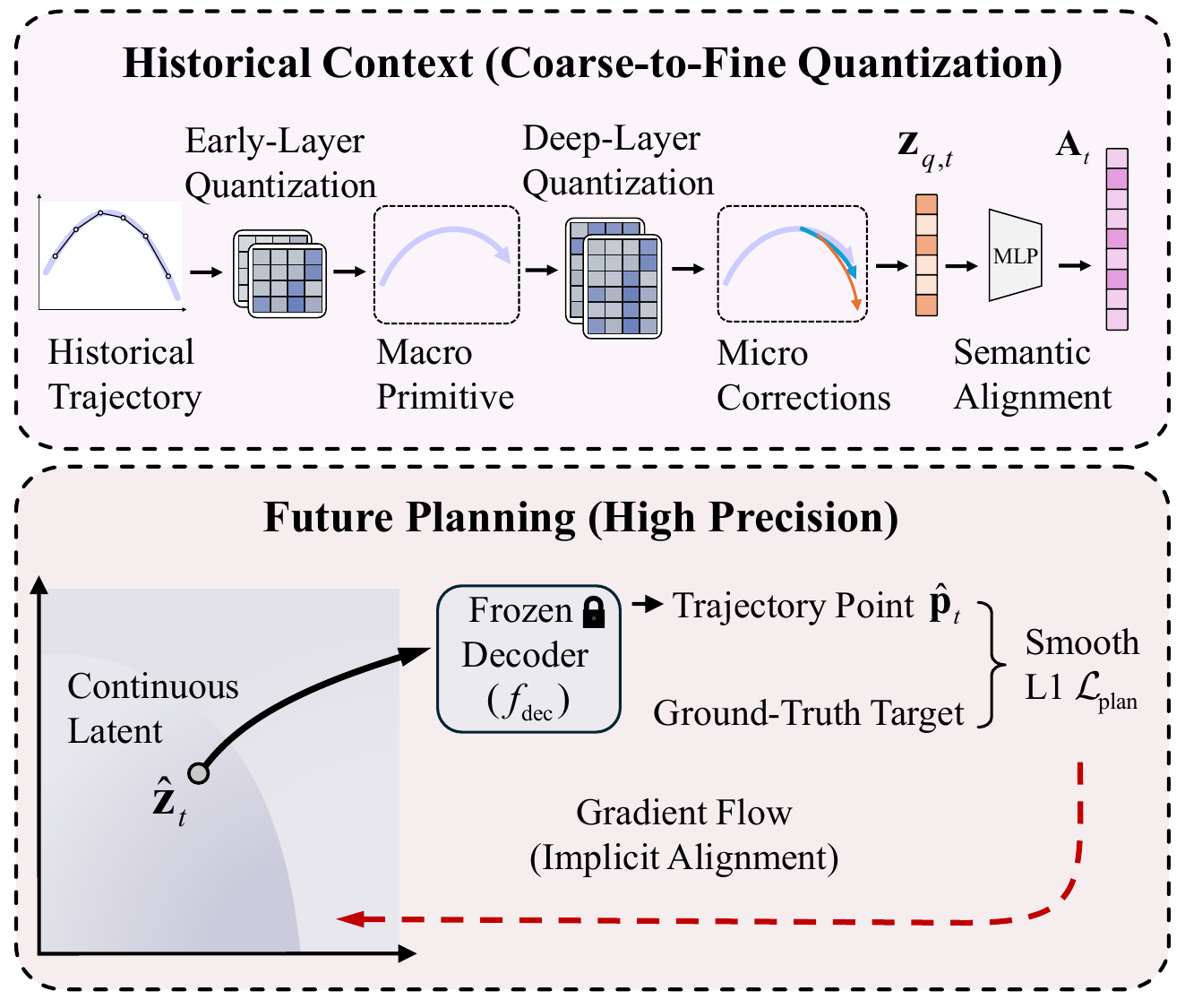} 
\caption{
\textbf{Asymmetric action representations.} We employ discrete quantization as an information bottleneck to structure historical context in a coarse-to-fine manner, while predicting continuous latents for the future to prevent quantization errors and ensure high-precision trajectory decoding.
}
\label{fig:discrete to conti}
\vspace{-1.5em}
\end{figure}

\subsubsection{Latent-Aligned Action Generation}
\label{sec:latent_aligned_action}

The core of LaPla is a trajectory prediction head that operates entirely in the continuous latent space of the frozen prior decoder.

\textbf{Parallel Prefill and Latent Projection.}
All \(T\) future action queries of \(\mathbf{P}\) are processed in a single forward pass. Let \(\mathbf{h}_t \in \mathbb{R}^{d_{\text{vlm}}}\) be the final-layer hidden state of \(\texttt{<act\_t>}\). A shared MLP projection head \(\phi_{\text{fut}}\) transforms each \(\mathbf{h}_t\) into a latent vector that lies in the input space of the frozen decoder:
\begin{equation}
\hat{\mathbf{z}}_t = \phi_{\text{fut}}(\mathbf{h}_t) \in \mathbb{R}^{d_z}, \qquad t \in \{0, \dots, T-1\},
\label{eq:latent proj}
\end{equation}
where \(d_z\) matches the decoder's input dimension.

\textbf{Trajectory decoding.}
To capture the multimodal nature of real-world driving, we generate diverse trajectory candidates using learned perturbations. Specifically, we maintain a set of $N$ learnable basis vectors $\mathbf{b}^{(n)} \in \mathbb{R}^{d_z}$ and corresponding scales $\sigma^{(n)} \in \mathbb{R}$, $n = 1, \dots, N$. 
At each future timestep $t$, the perturbed code for variant $n$ is defined as:
\begin{equation}
\hat{\mathbf{z}}_t^{(n)} = \hat{\mathbf{z}}_t + \sigma^{(n)} \mathbf{b}^{(n)} + 0.1\,\bigl( \hat{\mathbf{z}}_t \odot \mathbf{b}^{(n)} \bigr), \quad t \in \{0, \dots, T-1\},
\end{equation}
where $\odot$ denotes element-wise multiplication. 
The same perturbation is broadcast across all timesteps to maintain temporal consistency. 
The \(n\)-th set of perturbed latents is then fed into the frozen VQ‑VAE decoder to obtain the predicted displacement and heading at each future step:
\begin{equation}
[\hat{\mathbf{d}}_t^{(n)}; \hat{\theta}_t^{(n)}] = f_{\text{dec}}\bigl(\hat{\mathbf{z}}_t^{(n)}\bigr),    
\end{equation}
where \(\hat{\mathbf{d}}_t^{(n)}\) and \(\hat{\theta}_t^{(n)}\) denote the decoded displacement and heading, respectively. We retain \(\hat{\mathbf{d}}_t^{(n)}\) and reconstruct the absolute waypoints via cumulative summation from the current ego position \(\mathbf{p}_0\):
\begin{equation}
\begin{aligned}
    \hat{\mathbf{p}}_t^{(n)} &= \mathbf{p}_0 + \sum_{\tau=0}^{t-1} \hat{\mathbf{d}}_\tau^{(n)}, \quad t \in \{1, \dots, T\}.
\end{aligned}
\label{eq:decoding_multimodal}
\end{equation}
In this way, a single forward pass of the VLA produces \(N\) diverse trajectory candidates that originate from the same set of semantically grounded representations.

\textbf{Implicit latent alignment.}
During training, we supervise the predicted waypoint against the ground-truth \(\mathbf{p}_t\) with a smooth L1 loss in the Best-of-N (BoN) formulation:
\begin{equation}
\mathcal{L}_{\text{plan}} = \min_{n \in \{1,\dots,N\}} \frac{1}{T} \sum_{t=1}^{T} \big\| \hat{\mathbf{p}}_t^{(n)} - \mathbf{p}_t \big\|_{\text{SmoothL1},\beta}.
\label{eq:plan_loss}
\end{equation}
Here $\beta$ denotes the threshold parameter. 
Through the frozen decoder, this loss drives the \textbf{projected latents \(\hat{\mathbf{z}}_t\) to align with the manifold region that produces correct trajectories}, effectively realizing latent-aligned planning without auxiliary latent-space supervision. The continuous nature of the latent prediction allows fine-grained adjustment, avoiding the precision loss resulting from discretization, as shown in Fig.~\ref{fig:discrete to conti}.

\subsubsection{VQA as an Auxiliary Task}
\label{sec:vqa_auxiliary}

We incorporate a multi-round visual question answering task to strengthen the visual backbone's semantic grounding and to optionally provide interpretable outputs. The VQA dialogue \(\mathbf{Q}\) is appended after the action placeholders and consists of \(M\) question-answer pairs \(\{(q_m, a_m)\}_{m=1}^M\). The model generates each answer auto-regressively, conditioned on all preceding tokens:
\begin{equation}
\mathcal{L}_{\text{vqa}} = -\sum_{m=1}^{M} \sum_{p=1}^{L_m} \log P_{\theta}\big( a_{m,p} \mid \mathbf{H}, q_{\leq m}, a_{<m}, a_{m,<p} \big).
\label{eq:vqa_loss}
\end{equation}
Here, $\mathbf{H}$ denotes the cached hidden states of the input segment preceding to VQA. $q_{\leq m}$ denotes questions up to the $m$-th QA round, $a_{<m}$ denotes previous answers, and $a_{m,<p}$ represents answer tokens in the $m$-th round preceding to the $p$-th position. $L_m$ is the number of tokens in the $m$-th answer. During training, we randomly extract VQA pairs from the annotated dataset, exposing the model to varying dialog contexts to enhance generalization capacity.

\subsubsection{Training Objectives}
\label{sec:training_objectives}

The unified VLA framework is trained end-to-end with a joint loss:
\begin{equation}
\mathcal{L}_{\text{vla}} = \mathcal{L}_{\text{plan}} + \lambda \mathcal{L}_{\text{vqa}},
\label{eq:total_loss}
\end{equation}
where \(\lambda\) balances the two terms.

\textbf{Trainable Components.} All parameters of the pretrained VQ-VAE (\(f_{\text{enc}}\) and \(f_{\text{dec}}\)) remain frozen throughout LaPla training. The Emu3 backbone is adapted via LoRA for parameter-efficient fine-tuning. The learnable parameters thus consist of LoRA adapters, action queries $\mathbf{P}$, the history projector \(\phi_{\text{hist}}\), perturbation parameters $\{\mathbf{b}^{(n)}\}^{N}_{n=1}$ and $\{\sigma^{(n)}\}^{N}_{n=1}$, and the future latent projection head \(\phi_{\text{fut}}\).

\begin{table*}[t]
  \caption{Open-loop planning performance on the nuScenes benchmark. ST-P3 metrics are computed by cumulative average and UniAD metrics are computed per timestep.}
  \label{tab:nuscenes_quantitative}
  \centering
  \setlength{\tabcolsep}{3pt}
  \begin{tabular}{lcccccccccccccccc}
    \toprule
    \multirow{3}{*}{\textbf{Model}} 
    & \multicolumn{8}{c}{\textbf{ST-P3 metrics}} 
    & \multicolumn{8}{c}{\textbf{UniAD metrics}} \\
    \cmidrule(lr){2-9} \cmidrule(lr){10-17}
    & \multicolumn{4}{c}{\textbf{L2 (m) ↓}} 
    & \multicolumn{4}{c}{\textbf{Collision (\%) ↓}} 
    & \multicolumn{4}{c}{\textbf{L2 (m) ↓}} 
    & \multicolumn{4}{c}{\textbf{Collision (\%) ↓}} \\
    \cmidrule(lr){2-5} \cmidrule(lr){6-9} \cmidrule(lr){10-13} \cmidrule(lr){14-17}
    & 1s & 2s & 3s & Avg. 
    & 1s & 2s & 3s & Avg. 
    & 1s & 2s & 3s & Avg. 
    & 1s & 2s & 3s & Avg.\\
    \midrule

    ST-P3 \cite{hu2022st} & 1.33 & 2.11 & 2.90 & 2.11 & 0.23 & 0.62 & 1.27 & 0.71 & -& -& -& -& -& -& -& -\\
    VAD \cite{jiang2023vad}  & 0.17 & 0.34 & 0.60 & 0.37 & 0.07 & 0.10 & 0.24 & 0.14  & -& -& -& -& -& -& -& - \\
    UniAD \cite{hu2023planning} & 0.44 & 0.67 & 0.96 & 0.69 & 0.04 & 0.08 & 0.23 & 0.12 & 0.48 & 0.96 & 1.65 & 1.03 & 0.05 & 0.17 & 0.71 & 0.31\\
    GenAD \cite{zheng2024genad} & 0.28 & 0.49 & 0.78 & 0.52 & 0.08 & 0.14 & 0.34 & 0.19 & 0.36 & 0.83 & 1.55 & 0.91 & 0.06 & 0.23 & 1.00 & 0.43 \\
    EMMA \cite{hwang2024emma}& 0.14 & 0.29 & 0.54 & 0.32 & -& -& -& - & -& -& -& -& -& -& -& - \\
    OpenEMMA \cite{xing2025openemma} & 1.45 & 3.21 & 3.76 & 2.81 & -& -& -& - & -& -& -& -& -& -& -& - \\
    OmniDrive \cite{wang2025omnidrive} & 0.14 & 0.29 & 0.55 & 0.33 & 0.00 & 0.13 & 0.78 & 0.30 & - & - & - & - & - & - & - & -\\
    OpenDriveVLA-3B \cite{zhou2026opendrivevla} & 0.14 & 0.30 & 0.55 & 0.33 & 0.02 & 0.07 & 0.22 & 0.10 & 0.19 & 0.58 & 1.24 & 0.67 & 0.02 & 0.18 & 0.70 & 0.30\\
    OpenDriveVLA-7B \cite{zhou2026opendrivevla} & 0.15 & 0.31 & 0.55 & 0.33 & 0.01 & 0.08 & 0.21 & 0.10 & 0.20 & 0.58 & 1.21 & 0.66 & 0.00 & 0.22 & 0.55 & 0.25\\
    AutoVLA (action only) \cite{zhou2026autovla} &0.22 & 0.39 &  0.61 &0.41 & 0.10 & 0.17 & 0.28 & 0.18 & 0.29 & 0.67 & 1.17 & 0.71 & 0.15 & 0.34 & 0.56 & 0.35    \\
     AutoVLA (w/ CoT) \cite{zhou2026autovla} & 0.21 & 0.38 & 0.60 & 0.40 & 0.13 & 0.18 & 0.28 & 0.20  & 0.28 & 0.66 & 1.16 & 0.70 & 0.14 & 0.25 & 0.53 & 0.31\\ 
     DrivePI (w/o ego status) \cite{liu2026drivepi} & 0.24 & 0.46 & 0.78  & 0.49 & 0.38 & 0.27 & 0.48 & 0.38 & - & - & - & - & - & - & - & -\\
     DrivePI (w/ ego status) \cite{liu2026drivepi} & 0.19 & 0.36 & 0.64  & 0.40 & 0.00 & 0.05 & 0.28 & 0.11 & - & - & - & - & - & - & - & -\\
     \midrule
    \rowcolor{lightpurple}
    
    LaPla (ours) & 0.29 & 0.37 & 0.52  & 0.39 & 0.21 & 0.27 & 0.43 & 0.30 & 0.35 & 0.49 & 0.98 & 0.60 & 0.27 & 0.39 & 0.90 &  0.52   \\
    \bottomrule
  \end{tabular}
\end{table*}

\section{Experiment}

\subsection{Experiment Setup}
We comprehensively evaluate our approach in both open-loop and closed-loop experiments. Benchmark tests are conducted on nuScenes~\cite{caesar2020nuscenes} and AlpaSim~\cite{alpasim_2025}, respectively.

\textbf{nuScenes.} Following the split of \cite{zhou2026autovla} on the nuScenes dataset ($\sim$30K frames in total, annotated at 2\,Hz), we evaluate open-loop performance via \textbf{L2 distance} ($\downarrow$) and \textbf{collision rate} ($\downarrow$). Additionally, we incorporate VQA annotations from \cite{wang2025omnidrive} to enhance reasoning capabilities.

\textbf{AlpaSim.} We employ the high-fidelity NVIDIA AlpaSim for closed-loop evaluation. 20K frames are collected for additional fine-tuning, with 60 representative scenarios ($\sim$20\,s at 10 FPS) used for testing. Metrics include \textbf{fault collision} ($\downarrow$), \textbf{off road} ($\downarrow$), \textbf{route completion} ($\uparrow$), \textbf{success rate} ($\uparrow$, completion without fault collision or off road), and \textbf{FPS} ($\uparrow$).

LaPla is trained on a distributed computational platform equipped with Huawei Ascend 910B3 NPUs (64G VRAM). Open-loop baseline metrics are quoted from prior works. We use BoN inference with the oracle selector following \cite{zhou2026autovla}, setting $N$ to 20. For closed-loop evaluation, baselines are reproduced locally. LaPla is trained without perturbation operations and BoN for deterministic execution. AlpaSim runs on four RTX-4090 GPUs (24G VRAM), communicating with the NPU-backed planner via gRPC.

\subsection{Open-Loop Evaluations}

\subsubsection{Comparative Studies}
As shown in \Cref{tab:nuscenes_quantitative}, LaPla achieves competitive performance against state-of-the-art methods, \textbf{reducing the UniAD average L2 error to 0.60\,m,} outperforming OpenDriveVLA-7B (0.66\,m) and AutoVLA (0.70\,m). Crucially, this advantage is highly pronounced in long-horizon planning (3\,s), where \textbf{LaPla achieves 0.98\,m compared to AutoVLA's 1.16\,m in the UniAD L2 error, and 0.52\,m compared to EMMA's 0.54\,m in the ST-P3 L2 error.} The reduction in long-term deviation highlights that our latent-aligned parallel-inference architecture effectively overcomes the severe cumulative errors inherent in the autoregressive frameworks. By predicting continuous latents via a frozen kinematic prior, LaPla eliminates quantization bottlenecks, enabling precise and consistent motion execution that aligns with tactical goals.

\begin{table*}[t]
  \caption{Ablation studies on the action decoding strategy and multi-task learning. Lat-Act: Latent Action, Froz-Dec: Frozen Decoder.}
  \label{tab:ablation on action decode}
  \centering
  \setlength{\tabcolsep}{3pt}
  \begin{tabular}{c|ccc|cccccccccccccccc}
    \toprule
    \multirow{3}{*}{\textbf{Model}} 
    & \multicolumn{3}{c|}{\multirow{3}{*}{\textbf{Configuration}}} & \multicolumn{8}{c}{\textbf{ST-P3 metrics}} 
    & \multicolumn{8}{c}{\textbf{UniAD metrics}} \\
    \cmidrule(lr){5-12} \cmidrule(lr){13-20}
    & & & & \multicolumn{4}{c}{\textbf{L2 (m) ↓}} 
    & \multicolumn{4}{c}{\textbf{Collision (\%) ↓}} 
    & \multicolumn{4}{c}{\textbf{L2 (m) ↓}} 
    & \multicolumn{4}{c}{\textbf{Collision (\%) ↓}} \\
    \cmidrule(lr){2-4} \cmidrule(lr){5-8} \cmidrule(lr){9-12} \cmidrule(lr){13-16} \cmidrule(lr){17-20}
    & Lat-Act & Froz-Dec & VQA  & 1s & 2s & 3s & Avg. 
    & 1s & 2s & 3s & Avg. & 1s & 2s & 3s & Avg. & 1s & 2s & 3s & Avg.\\
    \midrule

     $\mathcal{M}_1$  & $\times$ & $\times$ & \checkmark & 0.51 & 0.97 & 1.54 & 1.00 & 0.35 & 0.50 & 1.15 & 0.66 & 0.70  & 1.70 & 3.03  & 1.81 & 0.38 & 1.04 & 3.03 & 1.48\\
     $\mathcal{M}_2$  & \checkmark & $\times$ & \checkmark & 0.44 & 0.67 & 0.95 & 0.69 & 0.13 & 0.41 & 0.87& 0.47 & 0.57 & 1.11  & 1.71  & 1.13 & 0.20 & 0.91 & 2.02 & 1.04\\
     $\mathcal{M}_3$ & \checkmark & \checkmark  & \checkmark & 0.29 & 0.37 & 0.52  & 0.39 & 0.21 & 0.27 & 0.43 & 0.30 & 0.35 & 0.49 & 0.98 & 0.60 & 0.27 & 0.39 & 0.90 &  0.52  \\
    $\mathcal{M}^{\prime}_3$ & \checkmark & \checkmark & $\times$ & 0.30 & 0.40 & 0.54  & 0.41 & 0.27 & 0.26 & 0.43 & 0.32 & 0.36 & 0.53 & 0.99 & 0.63 & 0.27 & 0.33 & 0.92 &  0.50 \\
    \bottomrule
  \end{tabular}
  \vspace{-1.5em}
\end{table*}

\subsubsection{Ablation Studies}
We establish a baseline ($\mathcal{M}_1$) that regresses trajectories via an MLP-based action head. As shown in \Cref{tab:ablation on action decode}, without latent alignment, $\mathcal{M}_1$ yields high average $L_2$ errors (1.00\,m in ST-P3, 1.81\,m in UniAD) and collision rates (0.66\,\% and 1.48\,\%, respectively).
Introducing the pretrained VQ-VAE decoder as a kinematic prior ($\mathcal{M}_2$) noticeably drops overall errors, reducing the UniAD $L_2$ from 1.81\,m to 1.13\,m and the collision rate from 1.48\,\% to 1.04\,\%. This confirms that anchoring outputs to a pretrained latent manifold effectively regularizes predictions against implausible maneuvers.
Integrating the frozen decoder ($\mathcal{M}_3$, our full LaPla) further cuts the UniAD $L_2$ error to 0.60\,m and the collision rate to 0.52\,\%. This demonstrates that freezing the trajectory decoder during policy learning eliminates the optimization burden of shifting action spaces. 
Finally, removing VQA ($\mathcal{M}^{\prime}_3$) degrades overall performance, validating its effectiveness in internalizing planning-oriented reasoning.

\subsubsection{Qualitative Analysis}
\begin{figure*}
    \centering
    \includegraphics[width=0.8\linewidth]{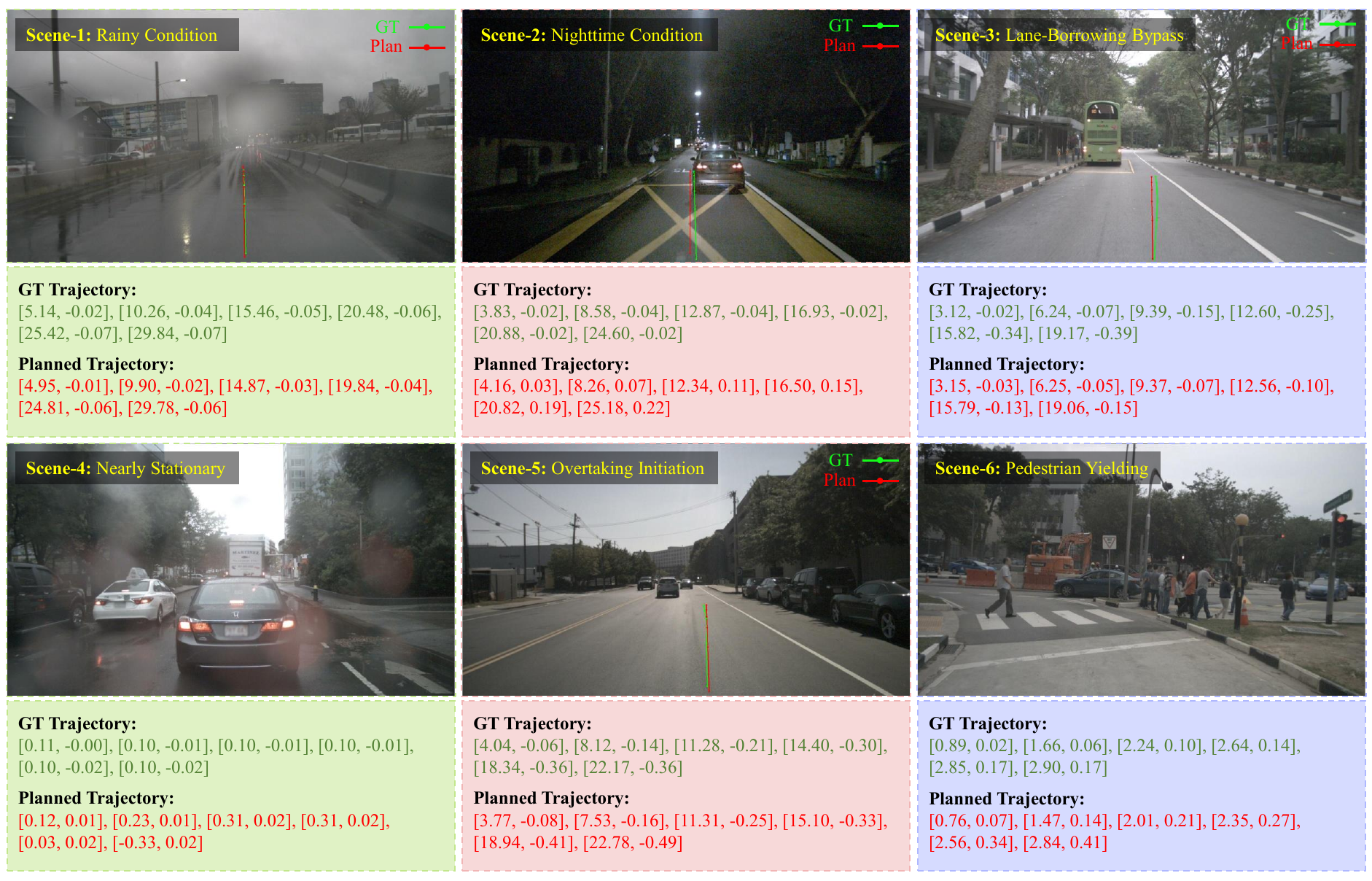}
    \vspace{-0.75em}
    \caption{\textbf{The open-loop planning performance of LaPla.} Visualizations span diverse driving conditions including rain (Scene~1), nighttime (Scene~2), lane-borrowing obstacle avoidance (Scene~3), near-stationary traffic (Scene~4), urban overtaking (Scene~5), and pedestrian yielding (Scene~6). 
LaPla preserves trajectory smoothness under perceptual degradation and consistently captures the correct maneuver intent in tactical scenarios.}
    \vspace{-0.5em}
 
    \label{fig:openloop qualitative}
\end{figure*}

Fig.~\ref{fig:openloop qualitative} visualizes representative cases across diverse conditions, revealing two consistent properties. 
First, \textbf{LaPla preserves trajectory smoothness under perceptual degradation.} In rainy and nighttime scenes (Scene~1,~2), it suppresses noise-induced jitter, maintaining temporal coherence despite visual occlusions or low-light noise.
Second, \textbf{in tactical lateral deviations, LaPla consistently captures the correct maneuver intent.} During the bypass of a stopped bus (Scene~3) and urban overtaking (Scene~5), the model initiates the expected lateral direction within the first second with monotonic progression, confirming that latent-aligned planning successfully grounds high-level decisions in the continuous action space.
In near-stationary traffic (Scene~4), mild directional ambiguity suggests under-represented creeping scenarios in the training data.
Overall, LaPla reliably translates multimodal context into physically coherent trajectories, highlighting the benefits of aligning semantic reasoning with a frozen trajectory prior.

\begin{figure*}
    \centering
    \includegraphics[width=0.8\linewidth]{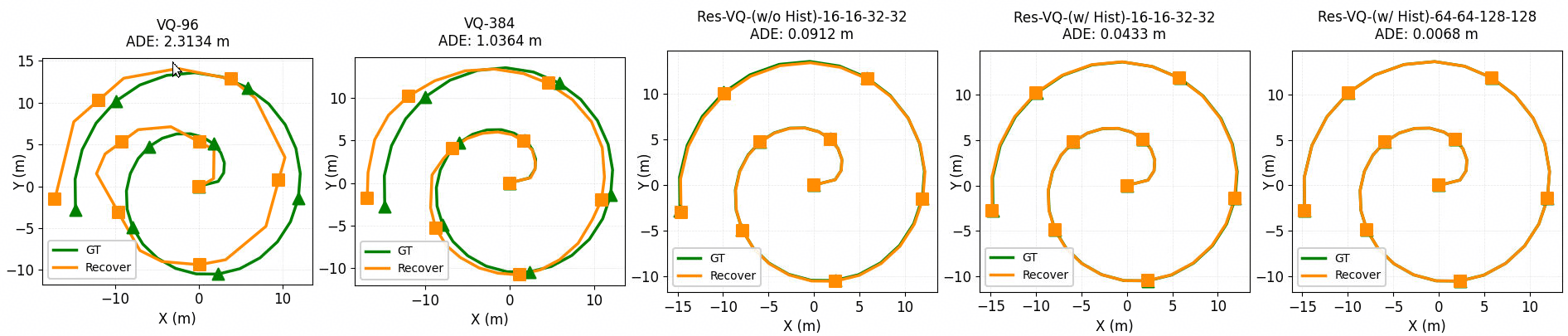}
    \vspace{-0.75em}
    \caption{\textbf{Reconstruction of a spiral trajectory generated by the kinematic bicycle model under test settings.} Notably, the reconstruction accuracy improves with the introduction of residual quantization and historical information.}
    \vspace{-1.5em}
 
    \label{fig:reconstruction qualitative}
\end{figure*}

\begin{table}[t]
  \caption{Action quantization with different codebook sizes and methods.}
  \label{tab:action quantization}
  \centering
  \resizebox{\linewidth}{!}{
  \begin{tabular}{l|c|ccc}
    \toprule
    \textbf{Method} & \textbf{Codebook Size} & \textbf{ADE (m) ↓} & \textbf{FDE (m) ↓} & \textbf{Usage (\%) $\uparrow$} \\

    \midrule

    \multirow{3}{*}{K-Disk}  & 96 & 0.2398 & 0.4622 & 100.00 \\
       & 192 & 0.1705 & 0.3242 & 99.48 \\
       & 384 & 0.1068 & 0.1962 & 96.35 \\
    \midrule

     \multirow{3}{*}{VQ}  & 96 & 0.3154 & 0.4438 & 100.00 \\
       & 192 & 0.1292  & 0.3465 & 100.00 \\
       & 384 & 0.1227  &0.2413 & 100.00  \\
    \midrule
    \multirow{3}{*}{\begin{tabular}[l]{@{}l@{}}Res-VQ\\(w/o Hist)\end{tabular}}   & \{16,16,32,32\} & 0.0184 & 0.0299 & 100.00 \\
       & \{32, 32, 64, 64\} & 0.0127  &  0.0208& 100.00 \\
       & \{64, 64, 128, 128\} & 0.0097  & 0.0162& 100.00  \\
        \midrule
    \multirow{3}{*}{\begin{tabular}[l]{@{}l@{}}Res-VQ\\(w/ Hist)\end{tabular}}   & \{16,16,32,32\} & 0.0084 & 0.0108 & 100.00 \\
       & \{32, 32, 64, 64\} & 0.0071 & 0.0086 &  100.00\\
       & \{64, 64, 128, 128\} & 0.0038 & 0.0046 &  100.00\\
    \bottomrule
  \end{tabular}
  }
  \vspace{-1.5em}
\end{table}

\subsubsection{Validation of Action Quantization}

Table \ref{tab:action quantization} demonstrates the superiority of our residual quantization. Res-VQ significantly outperforms flat codebook baselines like K-Disk and standard VQ. The $\{64, 64, 128, 128\}$ configuration achieves an ADE of $0.0038\,\text{m}$, a substantial reduction over the best K-Disk result, indicating that the hierarchical structure captures fine-grained action details more effectively. 
Furthermore, comparing Res-VQ (w/ Hist) and (w/o Hist) highlights the critical role of temporal context. Incorporating past trajectories reduces the ADE from $0.0097\,\text{m}$ to $0.0038\,\text{m}$ and the FDE from $0.0162\,\text{m}$ to $0.0046\,\text{m}$, suggesting the tokenizer successfully captures vehicle kinematic priors and temporal motion patterns.
Qualitatively, we evaluate the reconstruction of a complex spiral trajectory generated by a kinematic bicycle model. As shown in Fig.~\ref{fig:reconstruction qualitative}, the error progressively decreases with the residual mechanism, historical context, and increased codebook capacity. Ultimately, the proposed module ensures both high precision and temporal consistency.

\begin{table}[t]
  \caption{Closed-loop planning performance on AlpaSim. FC: Fault Collision, OR: Off Road, RC: Route Completion, SR: Success Rate. $\dagger$: Deployment with official checkpoint. } 
  \label{tab:closed-loop performance}
  \centering
  \resizebox{\linewidth}{!}{
  \begin{tabular}{lccccc}
    \toprule
    \textbf{Model} & \textbf{FC (\%) ↓} & \textbf{OR (\%) ↓} & \textbf{ RC (\%) $\uparrow$} & \textbf{SR (\%) $\uparrow$} & \textbf{FPS $\uparrow$}  \\

    \midrule

     AutoVLA (action only)  & 5.00 & 0.00 & 33.33   & 28.33 & 0.532 \\
     AutoVLA$^\dagger$ (w/CoT) & 1.67 & 0.00 & 15.00 & 15.00 & 0.215\\
     \rowcolor{lightpurple}
     LaPla (Ours)  & 11.67 & 0.00 & 71.67 & 61.67 & 1.582 \\
    \bottomrule
  \end{tabular}
  }
  \vspace{-1.5em}
\end{table}

\subsection{Closed-Loop Evaluations}
We evaluate our approach in the closed-loop simulation against two variants of AutoVLA. As shown in \Cref{tab:closed-loop performance}, \textbf{LaPla demonstrates a significant advancement in the success rate, achieving 61.67\,\% compared to the 28.33\,\% of AutoVLA (action only) and the 15.00\,\% of AutoVLA$^\dagger$ (w/CoT).} It is noted that LaPla better ensures closed-loop driving progress by maintaining a route completion rate of 71.67\,\%, while both variants of AutoVLA frequently become stuck due to the amplified errors accumulated from jagged planning trajectories. Our approach demonstrates a higher collision rate compared to AutoVLA owing to its more proactive driving behavior. In addition, \textbf{LaPla exhibits superior inference efficiency, achieving 1.582 FPS, which is nearly three times that of AutoVLA in fast inference mode (0.532 FPS).} This substantial speedup is attributed to our parallel latent prediction mechanism, while AutoVLA relies on the sequential decoding of discrete action tokens.

\section{Conclusion}
In this work, we introduced LaPla, a unified Vision-Language-Action framework bridging discrete semantic reasoning with continuous, physics-constrained motion required for autonomous driving.
LaPla leverages a frozen residual VQ-VAE decoder as a kinematic prior to predict continuous latent vectors in a single forward pass. This design effectively bypasses the quantization errors of discrete tokenization and the computational latency of autoregressive or diffusion-based generation.
Extensive evaluations on nuScenes and AlpaSim validate that LaPla achieves competitive open-loop accuracy and a significantly superior closed-loop success rate, offering a highly efficient, kinematically grounded solution for real-world VLA deployment.

\bibliography{sample}
\bibliographystyle{ieeetr}
\end{document}